\documentclass[11pt]{article}

\usepackage[preprint]{acl}

\usepackage{times}
\usepackage{latexsym}

\usepackage[T1]{fontenc}
\usepackage[utf8]{inputenc}
\usepackage{CJKutf8}

\usepackage{microtype}
\usepackage{inconsolata}
\usepackage{graphicx}
\usepackage{booktabs}
\usepackage{amsmath}
\usepackage{xcolor}
\usepackage{multirow}
\usepackage{makecell}
\usepackage{array}
\usepackage{fvextra}
\usepackage{mdframed}

\usepackage{adjustbox}

\definecolor{codebg}{RGB}{245,245,245}

\newmdenv[
  backgroundcolor=codebg,
  linecolor=gray!60,
  linewidth=0.5pt,
  innerleftmargin=3mm,
  innerrightmargin=3mm,
  innertopmargin=2mm,
  innerbottommargin=2mm,
]{codebox}

\newmdenv[
  linecolor=gray!55,
  linewidth=0.5pt,
  backgroundcolor=white,
  innerleftmargin=3mm,
  innerrightmargin=3mm,
  innertopmargin=2.5mm,
  innerbottommargin=2.5mm,
  skipabove=8pt,
  skipbelow=8pt,
]{examplebox}

\newcommand{\JPN}[1]{\begin{CJK}{UTF8}{ipxm}#1\end{CJK}}
\newcommand{\benchname}{JOR-Bench}

\title{JOR-Bench: Japanese Operations Research Benchmarks for \\Large Language Models}

\author{Yuu Jinnai \\
  CyberAgent \\
  Tokyo, Japan \\
  \texttt{jinnai\_yu@cyberagent.co.jp}}

\begin{document}
\maketitle

\begin{abstract}
We present \benchname{}, a collection of five Japanese-language benchmarks for evaluating the ability of large language models (LLMs) to formulate and solve operations research (OR) problems.
Each benchmark is a Japanese translation of an existing English benchmark: IndustryOR, MAMO Complex LP, NL4OPT, OptiBench, and OptMATH, covering 1,319 problems spanning linear programming, mixed-integer programming, non-linear programming, and combinatorial optimization.
\benchname{} is a solver-independent benchmark that can be used with any solver or programming language, and consists of pairs of Japanese problem statements and expected numerical answers.
We evaluate seven LLMs, including multilingual general-purpose models and Japanese-specialized models, on both the original English and the new Japanese versions, and compare performance across languages.
For the main evaluation, we standardize execution with the Python interface to OR-Tools to make model outputs comparable and reproducible with open-source software.
Our results show that OR formulation ability is largely language-neutral for strong multilingual models; the overall average accuracy difference between English and Japanese is only $-0.3$ pp.
Yet error analysis reveals subtle cross-lingual differences, including a pragmatic disambiguation failure in some domains that causes models to output decision-variable values instead of the objective value when the prompt is in Japanese.
\end{abstract}

%
%
\section{Introduction}
\label{sec:intro}

Operations research (OR) is concerned with mathematically optimizing decisions in complex systems~\cite{hillier2005introduction}.
Formulating a real-world problem as a mathematical program is a cognitively demanding task that has traditionally required specialized expertise.
With the rapid improvement of large language models (LLMs) in mathematical and code-generation tasks, there is growing interest in whether LLMs can automate OR problem formulation~\cite{nl4opt}.

Existing benchmarks for evaluating LLM-based OR formulation are predominantly English-language resources~\cite{yang2025optibench,lu2025optmath,orlm2025}.
This creates a gap for languages other than English: practitioners who work in non-English environments cannot reliably know whether an LLM will handle their natively phrased problems as well as English-phrased equivalents.
It is also of interest to the research community to understand whether LLMs' mathematical formulation ability is language-neutral or whether there are systematic cross-lingual differences.

In this paper we introduce \benchname{}, a collection of five Japanese-language benchmarks for evaluating LLMs' OR formulation ability derived from translating existing English benchmarks.
We conduct a cross-lingual evaluation of seven LLMs, including multilingual general-purpose models and Japanese-specialized models, on both the original English and the new Japanese versions of the benchmarks.
We find that overall OR formulation ability is largely language-neutral for strong multilingual models, with an average accuracy difference of only $-0.3$ pp between English and Japanese.

%
%
\section{\benchname{}: Japanese OR Benchmarks}
\label{sec:benchmarks}

\subsection{Source Benchmarks}

\benchname{} translates five existing English OR benchmarks into Japanese.

\textbf{IndustryOR}~\cite{orlm2025} contains 100 real-world industrial optimization
problems drawn from manufacturing, logistics, and supply chain management.
Problems typically involve multi-period planning with several decision variables
and realistic constraints.

\textbf{MAMO Complex LP}~\cite{huang-etal-2025-llms} consists of 203 linear programming problems collected and reviewed through a combination of human annotation and LLM-assisted quality checks.
All problems can be solved with a standard LP solver.

\textbf{NL4OPT}~\cite{nl4opt} is derived from the NL4OPT competition held in 2022.
The dataset contains 245 small LP problems expressed in English.
Problems typically involve two to three decision variables and a handful of linear constraints.

\textbf{OptiBench}~\cite{yang2025optibench} covers a wide variety of OR problems, including nonlinear programming problems gathered from textbooks by human workers.

\textbf{OptMATH}~\cite{lu2025optmath} is a synthetic dataset targeting hard combinatorial optimization and MIP problems, including job-shop scheduling, facility dispersion, and aircraft landing scheduling.

\subsection{Translation Methodology}

We translate all five English datasets into Japanese using LLM-based translation followed by human post-editing.
Because many source problems contain Markdown tables, structured data, and mathematical notation, preserving the input format is as important as translating the natural-language text.
We use gpt-oss-120b for translation because, in preliminary checks, it preserved tables and mixed-format problem statements more reliably than standard machine translation models we tested.
The first author then post-edited the translations to check terminology, numerical consistency, and faithfulness to the original English problem statements.

\begin{table*}
\centering
\adjustbox{max width=\linewidth}{
\begin{tabular}{p{0.45\linewidth}p{0.45\linewidth}p{0.1\linewidth}}
\toprule
\textbf{Japanese} & \textbf{English} & \textbf{Answer} \\
\midrule
\JPN{ある会計事務所は、パートタイム従業員とフルタイム従業員を雇用しています。フルタイム従業員は1シフトにつき8時間働き、パートタイム従業員は1シフトにつき4時間働きます。さらに、フルタイム従業員はシフトごとに300ドル、パートタイム従業員はシフトごとに100ドルを支払われます。現在、会計事務所には500時間の労働時間を必要とするプロジェクトがあります。予算が15,000ドルの場合、合計労働者数を最小にするために、各タイプの労働者を何人ずつスケジュールすべきでしょうか。} & An accounting firm employs part time workers and full time workers. Full time workers work 8 hours per shift while part time workers work 4 hours per shift. In addition, full time workers are paid \$300 per shift while part time workers are paid \$100 per shift. Currently, the accounting firm has a project requiring 500 hours of labor. If the firm has a budget of \$15000, how many of each type of worker should be scheduled to minimize the total number of workers. & 100 \\
\bottomrule
\end{tabular}
}
\caption{Example entry from \benchname{}.}
\label{tab:example}
\end{table*}

\subsection{Evaluation Protocol}

Table~\ref{tab:example} shows an example entry from \benchname{}.
Each entry pairs a Japanese problem description with the expected numerical answer.
The benchmark is therefore solver-independent: it can evaluate any LLM-generated solution as long as the final answer can be compared numerically.

For the experiments in this paper, we use the following execution-based protocol.
We prompt each model to generate a standalone Python script that formulates and solves the given optimization problem (see Appendix~\ref{apd:prompts}).
The model is required to surround the Python code with a fenced code block (triple backticks), which is a standard Markdown feature.
If such a code block is not found, we mark the problem as a ``no-code'' failure.
If a code block is found, we extract the code and attempt to execute it in a sandboxed Python environment with OR-Tools installed.
The sandbox enforces a maximum execution time of 300 seconds per problem and a 32GB memory limit.
If execution exceeds either limit, we mark the problem as a ``timeout'' or ``memory limit exceeded'' failure.
If the output of the run matches the expected answer within a tolerance of $10^{-6}$, we mark the problem as correct; otherwise, it is marked as incorrect.
Models occasionally generate code that uses only the Python standard library without importing OR-Tools.
We allow this behavior as long as the output is correct, because the benchmark is intended to measure problem-solving accuracy rather than strict compliance with a particular solver API.

%
%
\section{Experiments}
\label{sec:experiments}

The main experiments use OR-Tools as the execution backend. To evaluate backend sensitivity, we additionally report MAMO results on PuLP
and Pyomo in Appendix~\ref{apd:mamo-backend}.

\subsection{Models}

We evaluate seven LLMs spanning a range of sizes, training objectives, and
language specializations. All the models are reasoning models.
Table~\ref{tab:models} provides a summary.
Swallow is a Japanese-specialized model family continually pretrained from Qwen-3 series weights~\cite{fujii2024continual,ma2025building} to improve the Japanese language capabilities. We use the fine-tuned versions of Swallow as they demonstrate math and coding tasks roughly on par with Qwen-3 series models.
While Swallow is Japanese-specialized, it is trained to reason in English.
CAT-Thinking~\cite{jinnai2026costreasoningnonenglishlanguages} is a Japanese-specialized model further fine-tuned from Swallow-8B. 
CAT-Thinking is trained on a wide range of reasoning tasks including mathematical formulation and code generation and forced to generate reasoning trace in Japanese. Thus, at least on surface level, no English interferes the generation of the output when CAT-Thinking is prompted in Japanese. The performance of CAT-Thinking in the standard math and coding benchmarks is slightly lower than Swallow-8B.
Using these models allows us to evaluate the effect of Japanese specialization on OR formulation ability in two languages.
We additionally evaluate Nemotron, a Japanese-specialized model trained on a mixture of Japanese and English data~\cite{nvidia2025nvidianemotronnano2}, and GPT-OSS, a multilingual general-purpose model with a high competency~\cite{openai2024gptoss}.

\begin{table}[t]
\centering
\adjustbox{max width=\columnwidth}{
\begin{tabular}{llc}
\toprule
\textbf{Model} & \textbf{Full name} & \textbf{Params} \\
\midrule
Qwen3-8B     & Qwen/Qwen3-8B                          & 8B \\
Qwen3-32B    & Qwen/Qwen3-32B                         & 32B \\
Swallow-8B   & \makecell[l]{tokyotech-llm/\\Qwen3-Swallow-8B-RL-v0.2} \cite{ma2025building} & 8B \\
Swallow-32B  & \makecell[l]{tokyotech-llm/\\Qwen3-Swallow-32B-RL-v0.2} \cite{ma2025building}& 32B \\
CAT-Thinking & cyberagent/CAT-Thinking-8B \cite{jinnai2026costreasoningnonenglishlanguages} & 8B \\
Nemotron     & \makecell[l]{nvidia/NVIDIA-\\Nemotron-Nano-9B-v2-JP} \cite{nvidia2025nvidianemotronnano2} & 9B \\
GPT-OSS      & openai/gpt-oss-120b                    & 120B \\
\bottomrule
\end{tabular}
}
\caption{Models evaluated. All seven models are run on both the English and
  Japanese versions of all five benchmarks.}
\label{tab:models}
\end{table}

\subsection{Results}

Tables~\ref{tab:results_en} and~\ref{tab:results_ja} show accuracy (\%) on
English and Japanese benchmarks, respectively.

\begin{table}[t]
\centering
\adjustbox{max width=\columnwidth}{
\begin{tabular}{lrrrrr}
\toprule
\textbf{Model} & \textbf{Indus} & \textbf{MAMO} & \textbf{NL4OPT}
               & \textbf{OptiB} & \textbf{OptM} \\
\midrule
CAT-Thinking & \underline{77.0} & \textbf{85.2} & \textbf{81.6} & \textbf{77.9} & \underline{64.0} \\
GPT-OSS      & \textbf{81.0} & \underline{74.9} & 17.6 & 42.0 & \textbf{73.2} \\
Qwen3-8B     & 35.0 & 38.9 & 76.7 & 57.9 & 10.4 \\
Qwen3-32B    & 53.0 & 50.2 & \underline{80.0} & \underline{61.8} & 25.9 \\
Swallow-8B   & 27.0 & 12.3 & 45.7 & 33.1 & 7.3 \\
Swallow-32B  & 48.0 & 28.6 & 43.3 & 40.7 & 42.0 \\
Nemotron     & 9.0 & 4.9 & 4.1 & 3.0 & 7.0 \\
\bottomrule
\end{tabular}
}
\caption{Accuracy (\%) on \textbf{English} benchmarks.
  Bold = best; underline = second best per column.}
\label{tab:results_en}
\end{table}

\begin{table}[t]
\centering
\adjustbox{max width=\columnwidth}{
\begin{tabular}{lrrrrr}
\toprule
\textbf{Model} & \textbf{Indus} & \textbf{MAMO} & \textbf{NL4OPT}
               & \textbf{OptiB} & \textbf{OptM} \\
\midrule
CAT-Thinking & \textbf{81.8} & \textbf{85.0} & \underline{71.0} & \textbf{70.7} & \underline{66.3} \\
GPT-OSS      & \textbf{81.8} & \underline{73.5} & 21.2 & 46.4 & \textbf{71.7} \\
Qwen3-8B     & 48.5 & 34.5 & \textbf{78.4} & \underline{62.0} & 13.3 \\
Qwen3-32B    & \underline{55.6} & 50.0 & 70.6 & 58.8 & 19.5 \\
Swallow-8B   & 23.2 & 10.5 & 46.5 & 31.7 & 7.2 \\
Swallow-32B  & 50.5 & 37.5 & 42.9 & 41.7 & 45.1 \\
Nemotron     & 3.0 & 1.0 & 4.5 & 4.1 & 1.2 \\
\bottomrule
\end{tabular}
}
\caption{Accuracy (\%) on \textbf{Japanese} benchmarks.
  Bold = best; underline = second best per column.}
\label{tab:results_ja}
\end{table}

%
%
\section{Quantitative and Qualitative Analysis}
\label{sec:analysis}

Table~\ref{tab:delta} shows $\Delta_{\text{JA}} = \text{acc}_{\text{JA}} -
\text{acc}_{\text{EN}}$ for all seven models across the five benchmarks.

\begin{table}[t]
\centering
\adjustbox{max width=\columnwidth}{
\begin{tabular}{lrrrrrr}
\toprule
\textbf{Model} & \textbf{Indus} & \textbf{MAMO} & \textbf{NL4} & \textbf{OptiB}
               & \textbf{OptM} & \textbf{Avg} \\
\midrule
CAT-Thinking & $+4.8$  & $-0.2$ & $-10.6^{***}$ & $-7.1^{**}$ & $+2.2$ & $-2.2$ \\
GPT-OSS      & $+0.8$  & $-1.4$ & $+3.7$  & $+4.5^{**}$ & $-1.5$ & $+1.2$ \\
Qwen3-8B     & $+13.5^{*}$ & $-4.4$ & $+1.6$  & $+4.1$ & $+2.9$ & $+3.5$ \\
Qwen3-32B    & $+2.6$  & $-0.2$ & $-9.4^{**}$ & $-3.0$ & $-6.4$ & $-3.3$ \\
Swallow-8B   & $-3.8$  & $-1.8$ & $+0.8$  & $-1.3$ & $-0.1$ & $-1.2$ \\
Swallow-32B  & $+2.5$  & $+8.9^{*}$ & $-0.4$  & $+1.0$ & $+3.2$ & $+3.0$ \\
Nemotron     & $-6.0$  & $-3.9$ & $+0.4$  & $+1.2$ & $-5.8^{*}$ & $-2.8$ \\
\midrule
\textbf{Avg} & $+2.1$ & $-0.4$ & $-2.0$ & $-0.1$ & $-0.8$ & $-0.2$ \\
\bottomrule
\end{tabular}
}
\caption{$\Delta_{\text{JA}}$ (pp) = Japanese accuracy minus English accuracy.
  Significance from McNemar's test (continuity-corrected):
  $^{*}p{<}0.05$, $^{**}p{<}0.01$, $^{***}p{<}0.001$.
  Positive = Japanese is better; negative = English is better.}
\label{tab:delta}
\end{table}

\paragraph{Overall language neutrality.}
The grand average $\Delta_{\text{JA}}$ is $-0.3$ pp—statistically
indistinguishable from zero.
Most individual cells lie within $\pm 5$ pp.
This suggests that OR formulation ability, when assessed as code-generation
accuracy, is largely language-agnostic for models with sufficient multilingual
training.

\paragraph{Statistical significance.}
We apply McNemar's test (continuity-corrected) to per-problem binary outcomes
for each model--dataset pair.
Of 35 pairs, only 7 reach $p < 0.05$ (marked in Table~\ref{tab:delta}).
The four significant regressions concentrate on NL4OPT and OptiBench;
the three significant improvements span IndustryOR (Qwen3-8B, $p=0.026$),
MAMO (Swallow-32B, $p=0.030$), and OptiBench (GPT-OSS, $p=0.008$).
The remaining 28 cells are statistically indistinguishable from zero.

\paragraph{Model scale interacts with language.}
Scale helps, especially on hard problems, regardless of language.
Increasing model size within a family consistently improves accuracy, and the
gains are largest on the hardest benchmark.
On OptMATH (EN), Qwen3-32B more than doubles Qwen3-8B (25.9\% vs.\ 10.4\%) and
Swallow-32B improves over Swallow-8B nearly six-fold (42.0\% vs.\ 7.3\%).
Combinatorial optimization of this complexity appears to require substantial
model capacity regardless of language specialization.

That being said, the two model families for which we evaluate two sizes exhibit \emph{opposite}
scaling behavior with respect to language.
For Qwen3, the smaller 8B model improves in Japanese ($+3.5$ pp on average, the
best of any model) while the larger 32B model regresses ($-3.3$ pp), with a
$-9.4$ pp drop on NL4OPT and $-6.4$ pp on OptMATH.
For Swallow, the pattern reverses: the 8B model declines in Japanese
($-1.2$ pp) while the 32B model improves ($+3.2$ pp), including a $+8.9$ pp gain
on MAMO.
These crossovers show that cross-lingual transfer is not a fixed property of a
model family but depends on the interaction between scale and language
specialization.

\paragraph{MAMO declines for most models.}
MAMO shows the most consistent direction of negative transfer: six of the seven
models decline in Japanese, although the average drop is small ($-0.4$ pp)
because Swallow-32B improves by $+8.9$ pp.
We investigate this pattern in Section~\ref{sec:mamo}.

\subsection{MAMO: Japanese Decline for Most Models}
\label{sec:mamo}

We analyze error types on MAMO to understand why most models perform slightly
worse in Japanese.
Table~\ref{tab:mamo_errors} breaks down failures into our error taxonomy.
Backend sensitivity results with PuLP and Pyomo are reported separately in
Appendix~\ref{apd:mamo-backend}; this subsection focuses on OR-Tools-specific
error behavior.

\begin{table}[t]
\centering
\adjustbox{max width=\columnwidth}{
\begin{tabular}{lllrr}
\toprule
\textbf{Model} & \textbf{Lang} & \textbf{Correct} & \textbf{No-code} & \textbf{Import err} \\
\midrule
\multirow{2}{*}{GPT-OSS}   & EN & 74.9\% & 1.5\% & 18.7\% \\
                            & JA & 73.5\% & \textbf{4.0\%} & 18.5\% \\
\midrule
\multirow{2}{*}{Qwen3-8B}  & EN & 38.9\% & 20.7\% & 31.0\% \\
                            & JA & 34.5\% & \textbf{23.0\%} & 30.5\% \\
\midrule
\multirow{2}{*}{Swallow-8B} & EN & 12.3\% & 37.9\% & 18.7\% \\
                            & JA & 10.5\% & \textbf{47.5\%} & 14.0\% \\
\bottomrule
\end{tabular}
}
\caption{Error breakdown on MAMO for selected models.
  ``No-code'' = no fenced Python block found; ``Import err'' = deprecated
  OR-Tools API call. Bold highlights the increase in ``No-code'' errors
  from EN to JA.}
\label{tab:mamo_errors}
\end{table}

Two patterns are observed from the error breakdown.

\paragraph{Import errors are language-independent.}
The dominant error type is ``import error,'' caused by models attempting to use
the deprecated \texttt{ortools.graph.pywrapgraph} API.
This error occurs at nearly identical rates in English and Japanese for all
models (within 1 pp), confirming that it reflects a knowledge gap about
current API versions rather than a language effect.

\paragraph{No-code errors increase in Japanese.}
A consistent finding across the three models that decline is that \emph{no-code
failures increase in Japanese}.
Swallow-8B shows the most dramatic shift: 37.9\% in English versus 47.5\% in
Japanese.
This indicates that models are less likely to follow the implicit instruction
to respond with a Python code block when the prompt is in Japanese.
This is consistent with the broader finding in multilingual LLM research that
instruction-following behavior degrades in non-primary languages~\cite{dussolle-etal-2025-ifeval}.

\subsection{Pragmatic Disambiguation Failures}
\label{sec:error}

We examine why CAT-Thinking regresses substantially on NL4OPT ($-10.6$ pp)
and OptiBench ($-7.1$ pp) in Japanese.

\paragraph{NL4OPT error breakdown.}
In English, CAT-Thinking's primary failure mode is ``wrong answer'' (37/245
problems, 15.1\%).
In Japanese, wrong-answer failures increase to 56/245 (22.9\%),
accounting for the bulk of the regression.
Runtime errors also increase slightly (8 → 13).

A qualitative example illustrates the mechanism:

\begin{quote}
\textit{Problem (Japanese)}: An oil spill requires transporting ducks to shore.
A boat carries 10 ducks per trip (20 min/trip); a canoe carries 8 ducks (40
min/trip). The boat may make at most 12 trips; at least 60\% of all trips must
be by canoe. At least 300 ducks must be transported.
\JPN{ボートとカヌーをそれぞれ何回使うべきでしょうか？}\\
\textit{Expected answer}: 1160.0 (total time in minutes)\\
\textit{Model output}: \texttt{12 23} (boat trips and canoe trips)
\end{quote}

The model correctly solves the LP (12 boat trips, 23 canoe trips), but the
Japanese question asks \JPN{「何回使うべきか」} (``how many times should each
be used''), which it answers literally with decision variable values.
The English version of the same problem more strongly implies reporting the
optimal objective value (total time).
This is a language-specific disambiguation failure: in English, the implicit
pragmatics of ``minimize'' questions direct models to report the objective;
in Japanese, the explicit phrasing leads the model to a different answer.

\paragraph{OptiBench error breakdown.}
Similarly, OptiBench wrong-answer errors increase from 51 (8.4\%) in English
to 88 (14.5\%) in Japanese, while runtime errors remain approximately constant.
This supports the hypothesis that CAT-Thinking's domain adaptation encodes
stronger English-language pragmatics about what value to report as the answer.

\paragraph{Qwen3-32B: the same mechanism at larger scale.}
The same pragmatic disambiguation failure explains the significant Qwen3-32B
regression on NL4OPT ($-9.4$ pp, $p=0.005$).
Wrong-answer errors more than double in Japanese: 9.8\% (EN) $\to$ 18.4\%
(JA), while Qwen3-8B's wrong-answer rate is flat at 4.1\% in both languages.
The trend holds on OptiBench: Qwen3-32B wrong-answer errors rise from 6.9\%
to 13.2\% in Japanese, while Qwen3-8B stays below 1.3\%.
These findings unify the two significant NL4OPT regressions: both
CAT-Thinking ($-10.6$ pp, $p{<}0.001$) and Qwen3-32B ($-9.4$ pp, $p{=}0.005$)
are driven by wrong-answer inflation in Japanese, whereas the smaller Qwen3-8B
is unaffected.
Larger models, despite higher overall accuracy, appear more prone to treating
an explicit literal question (\JPN{何回使うべきか}, ``how many times should
each be used'') as the primary signal for what value to output.

\subsection{GPT-OSS on NL4OPT: A Format Mismatch}
\label{sec:gpt-nl4opt}

GPT-OSS scores only 17.6\% (EN) and 21.2\% (JA) on NL4OPT despite leading
on OptMATH.
The anomaly is language-independent: 78\% of failures in both languages
produce decision-variable output (e.g., \texttt{12 23} for boat and canoe
trips) rather than the expected objective value (\texttt{1160.0}).
Completions average $\sim$1{,}500 characters with no code truncation,
ruling out token-limit exhaustion as an explanation.

Per-problem EN/JA agreement is high ($\kappa = 0.70$), driven by 76\% of
problems failing consistently in both languages---the same problems, the same
failure mode.
A plausible explanation is a training-distribution mismatch: NL4OPT problems
frequently ask ``how many of each type should be used,'' phrasing that may
have been answered with variable counts in a distribution the model was
fine-tuned on.
The prompt explicitly requests the objective value (see Appendix~\ref{apd:prompts}),
but this instruction is overridden.
Unlike the failures in Section~\ref{sec:error}, this is a
language-independent benchmark-format issue.

\subsection{Per-Problem EN/JA Agreement}
\label{sec:agreement}

Aggregate accuracy differences can conceal per-problem churn if a model
solves \emph{different} problems correctly in each language.
We compute Cohen's $\kappa$ between per-problem binary correctness in
English and Japanese for each model--dataset pair.

The most striking result is Swallow-32B on OptMATH ($\kappa = {-0.07}$):
despite a near-neutral aggregate difference ($+3.2$ pp), only 19\% of
problems are correct in \emph{both} languages; 30\% are correct only in
Japanese and 23\% only in English.
Japanese and English thus act as complementary lenses on hard combinatorial
problems, unlocking different subsets for this model.
Qwen3-8B on OptMATH shows the same anti-correlated pattern ($\kappa = {-0.07}$,
EN+JA+ $= 0.6\%$): at the edge of model capability, language choice
switches solution and failure near-randomly.
Conversely, GPT-OSS on NL4OPT achieves high agreement ($\kappa = 0.70$) for
the wrong reason: 76\% of problems are consistently wrong in both languages
(Section~\ref{sec:gpt-nl4opt}).
These contrasts show that $\kappa$ reveals qualitatively different regimes
hidden beneath similar-looking accuracy numbers.

\subsection{Japanese-Specialized Models and Scale}

Among the Japanese-specialized models, scale is decisive.
Swallow-32B substantially outperforms Swallow-8B in Japanese on every benchmark
(e.g., MAMO 37.5\% vs.\ 10.5\%, OptMATH 45.1\% vs.\ 7.2\%, OptiBench 41.7\% vs.\
31.7\%), and unlike its smaller sibling it \emph{improves} in Japanese relative
to English ($+3.2$ pp on average).
Nemotron, the other Japanese-specialized model, remains weak at every scale of
comparison.
This indicates that Japanese specialization alone is not sufficient for OR
formulation: it must be paired with enough model capacity to reason about the
optimization problem.

\subsection{Qwen3 vs. Swallow: Effects of Japanese Continued Pretraining}
\label{sec:qwen-swallow}

Because Swallow models are built on Qwen3 with additional Japanese-focused
continued pretraining, we compare the two families directly at matched scales.
Table~\ref{tab:qwen_swallow_gap} reports the per-dataset accuracy gap
(Swallow minus Qwen3, in percentage points).

\begin{table}[t]
\centering
\small
\setlength{\tabcolsep}{3pt}
\begin{tabular}{lrrrr}
\toprule
\textbf{Dataset} & \textbf{EN 8B} & \textbf{EN 32B} & \textbf{JA 8B} & \textbf{JA 32B} \\
\midrule
IndustryOR & $-8.0$  & $-5.0$  & $-25.3$ & $-5.1$  \\
MAMO       & $-26.6$ & $-21.6$ & $-24.0$ & $-12.5$ \\
NL4OPT     & $-31.0$ & $-36.7$ & $-31.9$ & $-27.7$ \\
OptiBench  & $-24.8$ & $-21.1$ & $-30.3$ & $-17.1$ \\
OptMATH    & $-3.1$  & $+16.1$ & $-6.1$  & $+25.6$ \\
\midrule
\textbf{Avg} & \textbf{$-18.7$} & \textbf{$-13.7$} & \textbf{$-23.5$} & \textbf{$-7.4$} \\
\bottomrule
\end{tabular}
\caption{Family gap (pp) computed as Swallow accuracy minus Qwen3 accuracy
  at matched parameter scales. Positive values favor Swallow.}
\label{tab:qwen_swallow_gap}
\end{table}

\begin{figure}[t]
\centering
\includegraphics[width=\linewidth]{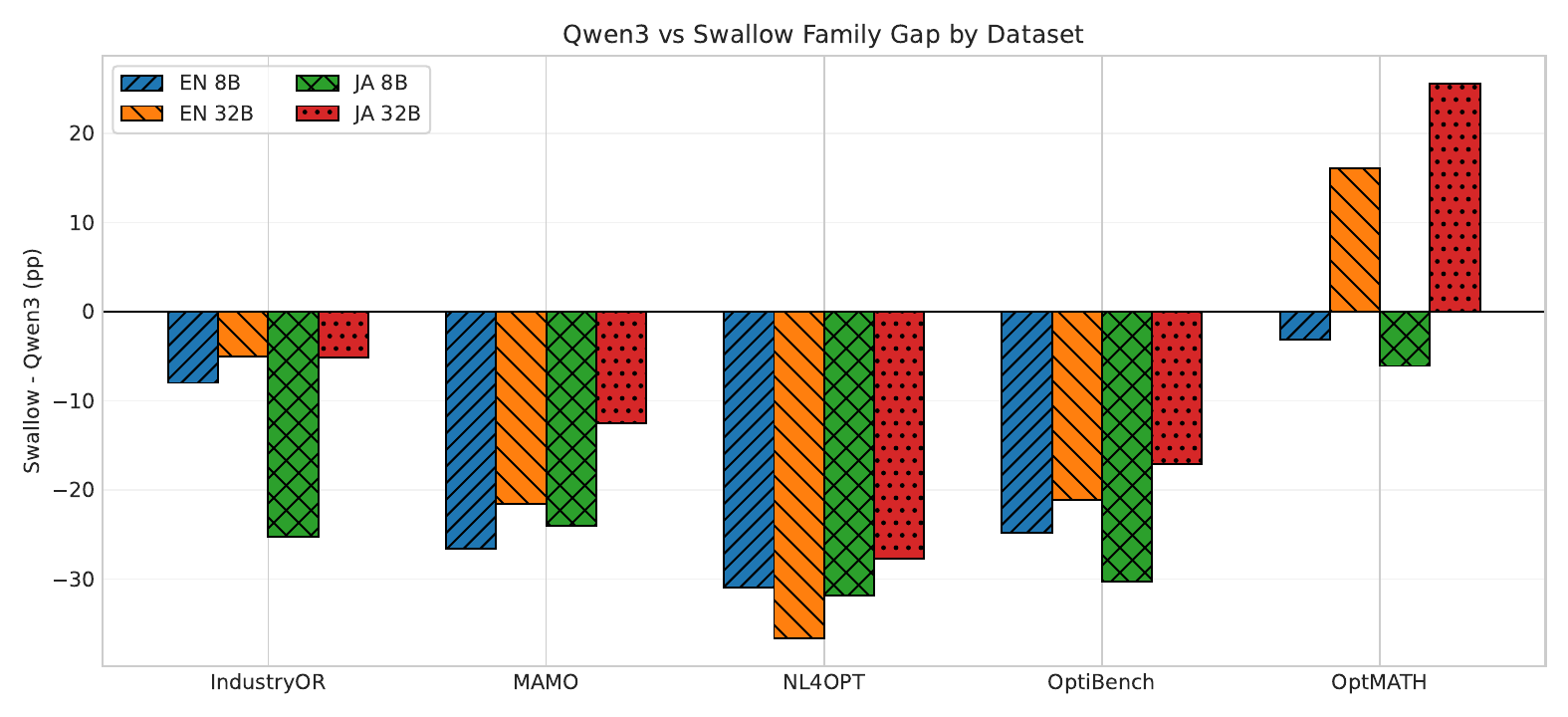}
\caption{Qwen3 vs. Swallow family gaps across datasets and languages.
  Hatches and color choices are designed to remain distinguishable in grayscale
  and for color-vision deficiencies.}
\label{fig:qwen_swallow_gap_main}
\end{figure}

Three findings are notable.
First, Qwen3 is consistently stronger on LP-heavy datasets (IndustryOR, MAMO,
NL4OPT, OptiBench) at both scales and in both languages.
Second, Swallow-32B substantially outperforms Qwen3-32B on OptMATH
($+16.1$ pp in English and $+25.6$ pp in Japanese), indicating that Japanese
continued pretraining can help on hard combinatorial optimization at larger
scale.
Third, the Japanese language setting narrows the average family gap at 32B
($-13.7$ to $-7.4$ pp), while widening it at 8B ($-18.7$ to $-23.5$ pp),
showing that the effect of language-focused continued pretraining depends on
model scale.

This family-level pattern is also backend-sensitive on MAMO
(Appendix~\ref{apd:mamo-backend}): under PuLP, Swallow outperforms Qwen3 at
both scales, while under Pyomo only Swallow-32B remains ahead.

\subsection{Dataset-Level Difficulty Analysis}

Table~\ref{tab:difficulty} summarizes the average accuracy across models for
each dataset and language.

\begin{table}[t]
\centering
\small
\begin{tabular}{lrrr}
\toprule
\textbf{Dataset} & \textbf{EN avg} & \textbf{JA avg} & \textbf{$\Delta$} \\
\midrule
IndustryOR  & 47.1\% & 49.2\% & $+2.1$ \\
MAMO        & 42.1\% & 41.7\% & $-0.4$ \\
NL4OPT      & 49.9\% & 47.9\% & $-2.0$ \\
OptiBench   & 45.2\% & 45.1\% & $-0.1$ \\
OptMATH     & 32.8\% & 32.0\% & $-0.8$ \\
\bottomrule
\end{tabular}
\caption{Average accuracy across all seven models per dataset.}
\label{tab:difficulty}
\end{table}

Averaged across models, English and Japanese accuracy are close on every
dataset (all $|\Delta| \le 2$ pp), supporting the overall language-neutrality
finding.
OptMATH consistently shows the lowest average accuracy (32.8\% EN, 32.0\% JA),
confirming it is the most challenging benchmark.
The remaining datasets cluster between 42\% and 50\%, indicating comparable
aggregate difficulty despite their different OR problem types.

%
%
\section{Conclusion}
\label{sec:conclusion}

We introduced \benchname{}, five Japanese-language OR benchmarks translated
from established English originals, together with a systematic evaluation of
seven LLMs across both language versions.

Our main finding is that OR formulation ability is largely language-neutral:
the grand average $\Delta_{\text{JA}}$ is only $-0.3$ pp across all
model--dataset pairs.
However, this aggregate masks important model-level and dataset-level patterns.
McNemar's tests confirm that only 7 of 35 model--dataset pairs differ
significantly; the remainder are statistically indistinguishable from zero.
A shared mechanism explains the two prominent NL4OPT regressions:
both CAT-Thinking ($-10.6$ pp, $p{<}0.001$) and Qwen3-32B ($-9.4$ pp,
$p{=}0.005$) show wrong-answer errors doubling in Japanese, as Japanese
phrasing elicits decision-variable output rather than the objective value,
while Qwen3-8B at 8B scale is unaffected.
GPT-OSS exhibits the same output-format issue on NL4OPT in \emph{both}
languages---a language-independent training-distribution mismatch.
Most models also decline on MAMO in Japanese not due to reasoning failures
but because they generate Python code blocks less reliably, a behavioral
instruction-following gap that only the largest Japanese-specialized model
(Swallow-32B) overcomes.
Additional MAMO robustness experiments with PuLP and Pyomo
(Appendix~\ref{apd:mamo-backend}) show that backend choice strongly affects
absolute accuracy, while the English--Japanese gap remains comparatively small.
Per-problem agreement (Cohen's $\kappa$) further reveals that on hard
combinatorial problems (OptMATH), Japanese and English unlock complementary
problem subsets: Swallow-32B solves 30\% of problems only in Japanese
and 23\% only in English, while the aggregate difference is just $+3.2$ pp.

These results underscore the value of Japanese OR benchmarks:
English performance alone does not predict Japanese performance,
and different models exhibit qualitatively different cross-lingual behaviors.
We release all data and evaluation code to facilitate multilingual OR research.

\section{Limitations}
This study has several limitations.
First, \benchname{} is a translation benchmark rather than a benchmark written
originally in Japanese.
It therefore may not fully capture the discourse patterns, terminology, and
implicit assumptions that appear in naturally occurring Japanese business or
engineering optimization problems. Therefore, \emph{the benchmark is Japanese-language but not culturally or contextually Japanese}.
Second, our error taxonomy is coarse; a more fine-grained categorization
(e.g., distinguishing constraint formulation errors from objective function
errors) would provide deeper insight.
Third, our execution-based evaluation is limited to Python solver interfaces
(OR-Tools in the main experiments, with PuLP and Pyomo in the MAMO robustness
study).
Preliminary experiments with MiniZinc and JuMP produced too few executable
formulations for most models to support a meaningful cross-lingual comparison;
broader multi-language solver evaluation is left for future work.

\JPN{

}
\appendix

\section{Solver Backend Robustness on MAMO}
\label{apd:mamo-backend}

Table~\ref{tab:mamo_backend} reports MAMO accuracy for additional execution
backends using the same 203 problems and five models.
Compared with OR-Tools, these PuLP and Pyomo runs show substantially lower
absolute accuracy overall.
Averaged over the five models, accuracy is 23.5\% (EN) and 23.0\% (JA) for
PuLP, and 16.3\% (EN) and 14.4\% (JA) for Pyomo.
Thus, backend choice materially changes absolute performance,
whereas the average cross-lingual gap remains relatively small
($-0.6$ pp for PuLP and $-1.9$ pp for Pyomo, JA minus EN).

\begin{table}[t]
\centering
\small
\setlength{\tabcolsep}{4pt}
\begin{tabular}{lrrrr}
\toprule
\textbf{Model} & \textbf{PuLP EN} & \textbf{PuLP JA} & \textbf{Pyomo EN} & \textbf{Pyomo JA} \\
\midrule
GPT-OSS      & 70.4 & 73.9 & 64.5 & 59.1 \\
Qwen3-8B     & 0.0  & 0.0  & 3.0  & 3.9  \\
Qwen3-32B    & 2.0  & 2.0  & 1.5  & 2.5  \\
Swallow-8B   & 13.8 & 11.8 & 0.0  & 0.5  \\
Swallow-32B  & 31.5 & 27.1 & 12.3 & 5.9  \\
\bottomrule
\end{tabular}
\caption{MAMO accuracy (\%) by backend and language for five models with
  available PuLP and Pyomo runs (all out of 203 problems).}
\label{tab:mamo_backend}
\end{table}

\begin{figure}[t]
\centering
\includegraphics[width=\linewidth]{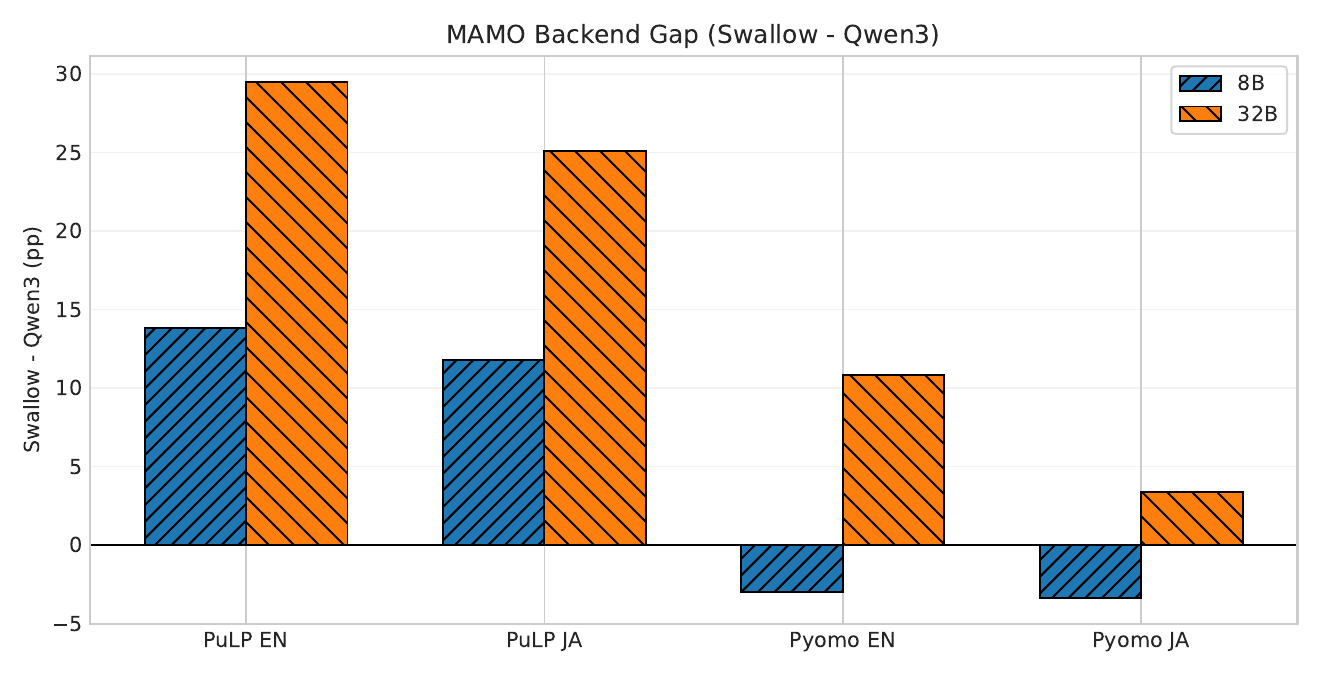}
\caption{MAMO backend sensitivity for the Swallow--Qwen3 family gap.
  The figure highlights that backend choice changes the absolute gap pattern,
  even when the model family and dataset are fixed.}
\label{fig:qwen_swallow_gap_mamo_backend}
\end{figure}

\section{Prompts}
\label{apd:prompts}

The prompts used for the experiments are provided below.
We use the same prompt template for all models and datasets, with only the
dataset-specific problem description, library name, and task language changing.
\begin{examplebox}
  Please develop a Python script that uses [[Library Name]] library to solve the mathematical problem provided below.
  
  \#\# Instructions
  
  1. Chain-of-Thought
  Before providing the code, explain the solution approach in [[Task Language]]. Your explanation should detail how to model the problem with [[Library Name]], specifically identifying the variables, constraints, and objective functions involved.
  2. Python Script Implementation
  Following the [[Task Language]] explanation, provide the Python script. Ensure the implementation adheres to these specific guidelines:
  
  - Code Quality: Write clean, high-quality code. Include helpful comments to ensure the logic is accessible to beginners.
  - Formatting: Enclose the script within a triple backtick code fence so it can be easily parsed.
  - Exact Output: The script's output must contain only the final numerical value or answer. Do not include any descriptive text (e.g., "Answer: 42") or additional formatting.
  - Mandatory [[Library Name]] Usage: You must solve the problem using [[Library Name]]. Do not hardcode the final answer, even for simple problems.
  - Standalone Execution: The script must be self-contained and ready to run.
  
  \#\# Problem to Solve

  [[Problem Statement Here]]
\end{examplebox}

\section{Reproducibility Statement}
\label{apd:reproducibility}

The experiments are conducted using the following software libraries and versions, as summarized in Table~\ref{tab:library}.
All the dataset used in the study are publicly available to facilitate reproducibility.
The experiments are run on NVIDIA A100 GPUs with 40GB or 80GB memory.

\begin{table}
  \centering
  \begin{tabular}{ll}
    \toprule
    Library & Version \\
    \midrule
    Python & 3.11.4 \\
    \midrule
    transformers & 4.57.6 \\
    vllm & 0.17.1 \\
    \midrule
    OR-Tools & 9.10.4067 \\
    PuLP & 2.4.0 \\
    Pyomo & 6.4.0 \\
    \bottomrule
  \end{tabular}
  \caption{Libraries used for the experiments.}
  \label{tab:library}
\end{table}

\end{document}